\documentclass[10pt,twocolumn,letterpaper]{article}

\usepackage[pagenumbers]{cvpr} %

\usepackage{times}
\usepackage{epsfig}
\usepackage{graphicx}
\usepackage{amsmath}
\usepackage{amssymb}
\usepackage[margin=1in]{geometry}
\usepackage{xcolor}
\usepackage[utf8]{inputenc}
\usepackage{hhline}
\usepackage{color, colortbl}
\usepackage{multirow}
\usepackage{times}
\usepackage{epsfig}
\usepackage{graphicx}
\usepackage{amsmath}
\usepackage{amssymb}
\usepackage{subcaption}
\usepackage[accsupp]{axessibility}
\usepackage{lipsum}
\usepackage{booktabs}
\usepackage{makecell}
\usepackage{url}

\usepackage[pagebackref,breaklinks,colorlinks]{hyperref}

\usepackage[capitalize]{cleveref}
\crefname{section}{Sec.}{Secs.}
\Crefname{section}{Section}{Sections}
\Crefname{table}{Table}{Tables}
\crefname{table}{Tab.}{Tabs.}

\def\cvprPaperID{1} %
\def\confName{CVPR}
\def\confYear{2023}

\hypersetup{
urlcolor=blue,
}
\begin{document}

\title{Consistency and Accuracy of CelebA Attribute Values}

\author{Haiyu Wu$^{1}$, Grace Bezold$^{1}$, Manuel G\"unther$^{2}$, Terrance Boult$^3$, \\Michael C. King$^{4}$, Kevin W. Bowyer$^{1}$\\
$^{1}$University of Notre Dame, $^2$University of Zurich, \\$^3$University of Colorado Colorado Springs, $^{4}$Florida Institute of Technology\\}

\maketitle

\begin{abstract}
We report the first systematic analysis of the experimental foundations of facial attribute classification. Two annotators independently assigning attribute values shows that only 12 of 40 common attributes are assigned values with $\geq 95\%$ consistency, and three (high cheekbones, pointed nose, oval face) have essentially random consistency.  Of 5,068 duplicate face appearances in CelebA, attributes have contradicting values on from 10 to 860 of the 5,068 duplicates. Manual audit of a subset of CelebA estimates error rates as high as 40\% for (no beard=false), even though the labeling consistency experiment indicates that no beard could be assigned with $\geq 95\%$ consistency. Selecting the mouth slightly open (MSO) for deeper analysis, we estimate the error rate for (MSO=true) at about 20\% and (MSO=false) at about 2\%. A corrected version of the MSO attribute values enables learning a model that achieves higher accuracy than previously reported for MSO.  Corrected values for CelebA MSO are available at \url{https://github.com/HaiyuWu/CelebAMSO}.

\end{abstract}

\section{Introduction}

\begin{figure}
    \centering
    \begin{subfigure}[b]{1\linewidth}
    \captionsetup[subfigure]{labelformat=empty}
        \includegraphics[width=\linewidth]{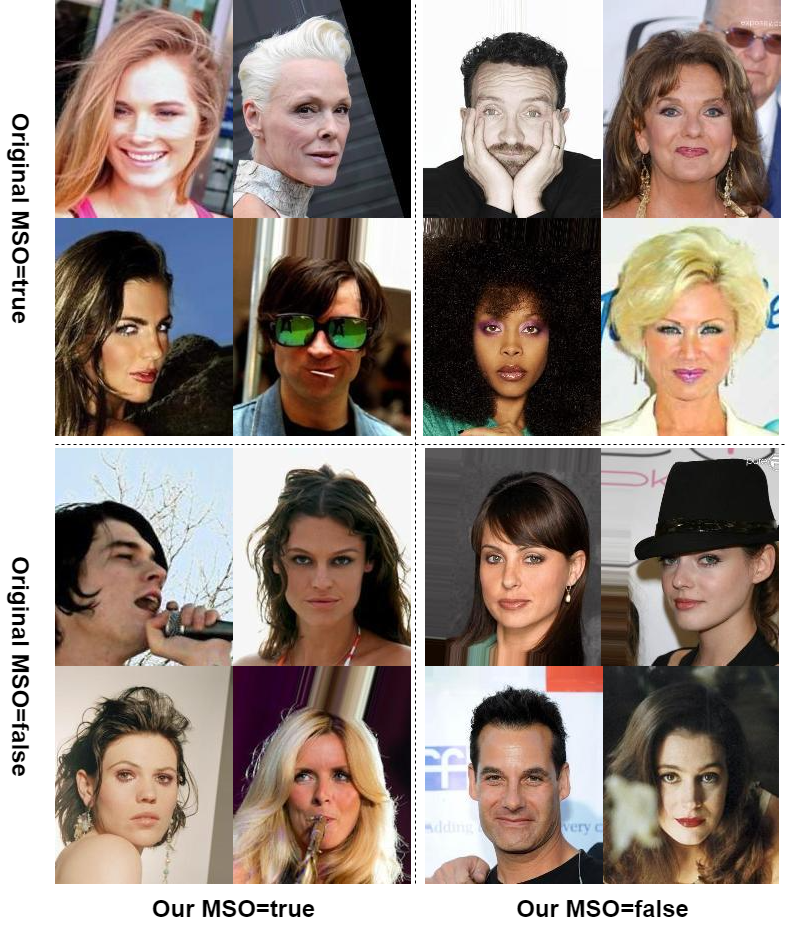}
    \end{subfigure}
   \caption{ {\it Which Set of Attribute Values Enables Learning a Better Model?} 
   Lower left quadrant contains images with original (MSO=false) corrected to (MSO=true);
   Upper right quadrant contains images with original (MSO=true) corrected to (MSO=false).
   }
\label{fig:teaser-figure}
\end{figure}

Facial attributes have potential uses in face matching/recognition \cite{berg2013poof, chan2017face, kumar2011describable, kumar2009attribute, manyam2011two, song2014exploiting}, face image retrieval \cite{li2015two, nguyen2018large}, re-identification~\cite{shi2015transferring, su2017multi, su2016deep}, training GANs~\cite{choi2018stargan, choi2020stargan, he2019attgan, li2021semantic} for generation of synthetic images, analyzing AI biases~\cite{xu2022comprehensive, wu2023logical} and other areas.
CelebA~\cite{CelebA_liu} is the largest and most widely used dataset in this research area.
However, recent papers have described cleaning the identity groups in CelebA~\cite{celeca_identity}, and suggested that the facial attribute annotations~\cite{gunther2017affact, thom2020facial} need some ``cleaning''.
This paper provides the first analysis of the consistency with which the commonly-used face attributes can be manually marked, and also of the quality of the attribute values distributed with the CelebA images.  We also propose an auditing workflow to clean existing annotations, and demonstrate that using a corrected set of attribute values enables learning a substantially different and more accurate model.
Contributions of this work include:
\begin{itemize}
\item
Analysis of independent manual annotation of the 40 commonly used face attributes shows that only 12 are labeled with $\geq$ 95\% consistency and 3 have random (50\%) consistency.
(See Section~\ref{section:consistency_subjectivity_accuracy_estimation}.)
\item 
For the 12 attributes that we determined can be labeled with $\geq$ 95\% consistency across annotators, we audit the attribute values provided with the CelebA images and find that (1) the error rate is often asymmetric between true / false, and (2) the error rate is as high as 40\% for some attribute values.
(See Section~\ref{section:consistency_subjectivity_accuracy_estimation}.)

\item
We propose a semi-automated workflow to clean existing annotations, and use it to create corrected MSO attribute values for CelebA.
In part of this correction/cleaning, we identify (1) a small number of images that are unusable and we propose should be dropped from CelebA, and (2) identify images for which true/false cannot be assigned to a particular attribute and so we propose an ``information not visible’’ value must be introduced.
 (See Sections~\ref{section:workflow} and \ref{section:definition_mso}.)
\item
Comparing models learned using the original MSO values versus our cleaned values, we show that the models are substantially different, and that our cleaned values enable a model that achieves state-of-the-art accuracy on MSO.
(See Section~\ref{section:network performance}.)
\end{itemize}

\section{Related work}

There is a large literature in facial attribute analysis, and several surveys give a broad coverage of the field
\cite{Arigbabu_survey_2015, Becerra-Riera_survey_2019, thom2020facial, Zheng_survey_2020}.
We cover only a few of the most relevant works here.

CelebA was introduced by Liu et al. \cite{CelebA_liu} in 2015 specifically to support research in deep learning for facial attributes.
CelebA has 202,599 images grouped into 10,177 identities.
Each image has 40 true/false attributes -- pointy nose, oval face, gray hair, wearing hat, etc. -- and five landmark locations.
CelebA also has a recommended subject-disjoint split into train, validation and test.
The creation of the attribute values is described only as  – “Each image in CelebA and LFWA is annotated with forty facial attributes and five key points by a professional labeling company” \cite{CelebA_liu}.
No description of how the attribute values are created, or estimate of their consistency or accuracy, is given.

Thom and Hand stated \cite{thom2020facial} that, ``CelebA and LFWA were the first (and only to date) large-scale datasets introduced for the problem of facial attribute recognition from images. Prior to CelebA and LFWA, no dataset labeled with attributes was large enough to effectively train deep neural networks.''
In number of identities and images, CelebA is substantially larger than LFWA, and is the most-used research dataset in this area. In addition, Thom and Hand speculate that noise in the attribute values may lead to an apparent plateau in research progress \cite{thom2020facial} – ``There is a recent plateau in facial attribute recognition performance, which could be due to poor labeling of data. … While crowdsourcing such tasks can be very useful and result in large quantities of reasonably labeled data, there are some tasks which may be consistently labeled incorrectly ...''.

The only paper we are aware of to discuss cleaning CelebA is %
\cite{celeca_identity}.
They deal specifically with errors in the identity groupings, and do {\it not} consider errors in the  attribute values. Compared to the original 202,599 images of 10,177 identities, their identity-cleaned version has 197,477 images of 9,996 identities.
In a simple manual check of 100 identities in CelebA, we found a few additional instances of identity errors in the identity-cleaned version of \cite{celeca_identity}.
Because our work is focused on attribute classification and not face matching, we start with the original CelebA rather than the version of \cite{celeca_identity}.

Terh{\"o}rst et al. \cite{maad-face} firstly estimated the quality of annotations in CelebA by letting three human evaluators manually evaluate randomly selected 50 positively-annotated and 50 negatively-annotated images for each attribute. They claimed that "Similar to LFW, there is a tendency that most of the wrong annotations are within the positives". In this paper, we provide a more statistically and systematically analysis on each attribute, and we, furthermore, provide a possible solution to clean the dataset.

\begin{figure*}[t]
\centering
    \begin{subfigure}[b]{1\linewidth}
    \captionsetup[subfigure]{labelformat=empty}
    \centering
        \begin{subfigure}[b]{0.72\linewidth}
            \begin{subfigure}[b]{0.32\linewidth}
            \centering
                \begin{subfigure}[b]{0.48\linewidth}
                    \includegraphics[width=\linewidth]{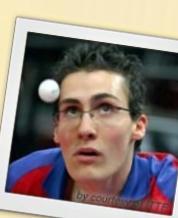}
                \end{subfigure}
                \begin{subfigure}[b]{0.48\linewidth}
                    \includegraphics[width=\linewidth]{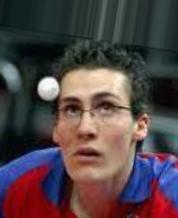}
                \end{subfigure}
            \end{subfigure}
            \begin{subfigure}[b]{0.32\linewidth}
            \centering
                \begin{subfigure}[b]{0.48\linewidth}
                    \includegraphics[width=\linewidth]{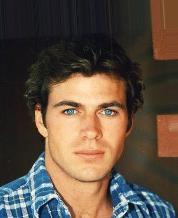}
                \end{subfigure}
                \begin{subfigure}[b]{0.48\linewidth}
                    \includegraphics[width=\linewidth]{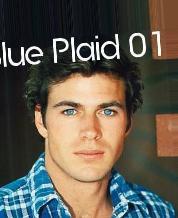}
                \end{subfigure}
            \end{subfigure}
            \begin{subfigure}[b]{0.32\linewidth}
            \centering
                \begin{subfigure}[b]{0.48\linewidth}
                    \includegraphics[width=\linewidth]{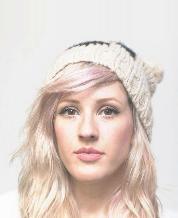}
                \end{subfigure}
                \begin{subfigure}[b]{0.48\linewidth}
                    \includegraphics[width=\linewidth]{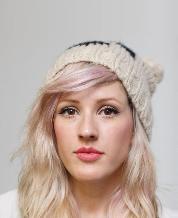}
                \end{subfigure}
            \end{subfigure}
        \end{subfigure}
        \begin{subfigure}[b]{0.23\linewidth}
        \centering
            \begin{subfigure}[b]{0.48\linewidth}
                \includegraphics[width=\linewidth]{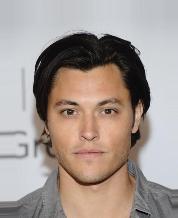}
            \end{subfigure}
            \begin{subfigure}[b]{0.48\linewidth}
                \includegraphics[width=\linewidth]{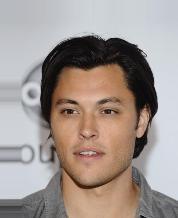}
            \end{subfigure}
        \end{subfigure}
    \end{subfigure}
   \caption{Example duplicate and near-duplicate pairs. The first three pairs have duplicate facial appearance, although the overall images have some differences from editing.  The fourth (rightmost) pair is only a near-duplicate, with highly similar but different face appearance; e.g., mouth closed in the left image, open in right image.}
\label{fig:duplicate_pair_samples}
\end{figure*}

Motivated by comments in \cite{gunther2017affact, thom2020facial} on the importance of correct labels for machine learning and errors encountered in CelebA attribute values, we present the first detailed analysis of the accuracy of CelebA attribute values.
We create a cleaned version of the MSO values, and perform experiments to assess the impact of the original versus cleaned MSO attribute, using AFFACT \cite{gunther2017affact}, MOON \cite{rudd2016moon}, DenseNet \cite{huang2017densely} and ResNet \cite{he2016deep}.
We show that the cleaned attribute values result in learning a more coherent model that achieves higher accuracy.

\vspace{-2mm}
\section{Accuracy of Attributes In Training Data}
\label{section:consistency_subjectivity_accuracy_estimation}

We first examine the general consistency of manual annotations of face attributes.
Then we use duplicate faces in CelebA to examine the consistency of the attribute values distributed with CelebA.
Then we manually audit a sample of CelebA to estimate the accuracy of its attribute values.

\subsection{Consistency of Manual Annotations?}
\begin{table}[t]
\small
\centering
\begin{tabular}{|l|l|l|l|}
\hline
Attribute         & $N_{d}$ & Attribute                                 & $N_{d}$                        \\ \hline
\rowcolor[HTML]{FCE4D6} 
High Cheekbones & 512  & \cellcolor[HTML]{FFE699}Smiling           & \cellcolor[HTML]{FFE699}141 \\ \hline
\rowcolor[HTML]{FCE4D6} 
Pointy Nose       & 490 & \cellcolor[HTML]{FFE699}Wearing Lipstick     & \cellcolor[HTML]{FFE699}133 \\ \hline
\rowcolor[HTML]{FCE4D6} 
Oval Face       & 466  & \cellcolor[HTML]{FFE699}Blurry            & \cellcolor[HTML]{FFE699}118 \\ \hline
\rowcolor[HTML]{FCE4D6} 
Arched Eyebrows   & 406 & \cellcolor[HTML]{FFE699}5 o Clock Shadow   & \cellcolor[HTML]{FFE699}110 \\ \hline
\rowcolor[HTML]{FCE4D6} 
Narrow Eyes     & 394  & \cellcolor[HTML]{FFE699}Chubby            & \cellcolor[HTML]{FFE699}110 \\ \hline
\rowcolor[HTML]{FCE4D6} 
Attractive       & 377  & \cellcolor[HTML]{FFE699}Sideburns         & \cellcolor[HTML]{FFE699}105 \\ \hline
\rowcolor[HTML]{FCE4D6} 
Straight Hair   & 369  & \cellcolor[HTML]{FFE699}Blond Hair       & \cellcolor[HTML]{FFE699}101 \\ \hline
\rowcolor[HTML]{FCE4D6} 
Wavy Hair       & 299  & \cellcolor[HTML]{FFE699}Double Chin      & \cellcolor[HTML]{FFE699}58  \\ \hline
\rowcolor[HTML]{FCE4D6} 
Big Nose        & 286  & \cellcolor[HTML]{C6E0B4}Wearing Earrings & \cellcolor[HTML]{C6E0B4}50  \\ \hline
\rowcolor[HTML]{FCE4D6} 
Bags Under Eyes  & 284 & \cellcolor[HTML]{C6E0B4}Wearing Necklace     & \cellcolor[HTML]{C6E0B4}49  \\ \hline
\rowcolor[HTML]{FCE4D6} 
Brown Hair      & 284  & \cellcolor[HTML]{C6E0B4}Gray Hair        & \cellcolor[HTML]{C6E0B4}37  \\ \hline
\rowcolor[HTML]{FCE4D6} 
Pale Skin       & 237  & \cellcolor[HTML]{C6E0B4}No Beard         & \cellcolor[HTML]{C6E0B4}34  \\ \hline
\rowcolor[HTML]{FCE4D6} 
Black Hair      & 232  & \cellcolor[HTML]{C6E0B4}Wearing Necktie  & \cellcolor[HTML]{C6E0B4}27  \\ \hline
\rowcolor[HTML]{FCE4D6} 
Rosy Cheeks     & 217  & \cellcolor[HTML]{C6E0B4}Male              & \cellcolor[HTML]{C6E0B4}23  \\ \hline
\rowcolor[HTML]{FCE4D6} 
Big Lips        & 206  & \cellcolor[HTML]{C6E0B4}Mustache          & \cellcolor[HTML]{C6E0B4}22  \\ \hline
\rowcolor[HTML]{FCE4D6} 
Heavy Makeup    & 181  & \cellcolor[HTML]{C6E0B4}Bald              & \cellcolor[HTML]{C6E0B4}18  \\ \hline
\rowcolor[HTML]{FCE4D6} 
Receding Hairline & 178 & \cellcolor[HTML]{C6E0B4}MSO & \cellcolor[HTML]{C6E0B4}17  \\ \hline
\rowcolor[HTML]{FCE4D6} 
Young            & 169  & \cellcolor[HTML]{C6E0B4}Goatee            & \cellcolor[HTML]{C6E0B4}16  \\ \hline
\rowcolor[HTML]{FCE4D6} 
Bushy Eyebrows  & 158  & \cellcolor[HTML]{C6E0B4}Wearing Hat      & \cellcolor[HTML]{C6E0B4}14  \\ \hline
\rowcolor[HTML]{FCE4D6} 
Bangs            & 150  & \cellcolor[HTML]{C6E0B4}Eyeglasses        & \cellcolor[HTML]{C6E0B4}3   \\ \hline
\end{tabular}
\caption{
Number of differences ($N_{d}$) between manual annotations of 1,000 images.
$N_{d}$ grouped into three subjectivity levels, from pink (high) to green (low).
}
\label{table:consistency_1000}
\end{table}

To estimate the level of consistency that can be expected in manual annotations of commonly used face attributes, two annotators independently assigned values for each of 40 attributes of 1,000 images.
The images were randomly selected from CelebA and should be representative of web-scraped, in-the-wild celebrity images.
The annotators viewed the cropped, normalized face images, with no knowledge of the other annotator's results.

Table~\ref{table:consistency_1000} lists, for each attribute, the number of images for which the two annotations disagree.
The least disagreement was for eyeglasses, at just 3 images.
The 3 images present an ``edge case'' for the attribute definition. 
There are eyeglasses in the image, but positioned on the person's head rather than in front of their eyes.
One annotator marked these as eyeglasses=true and the other marked these as eyeglasses=false.
Similarly, the 14 disagreements on the wearing\_hat attribute were primarily instances of a literal hat versus a more general head covering (scarf, visor, ...).
Inconsistencies arising from this type of ambiguity in the attribute definition should be reduced if annotators are provided a sufficiently detailed definition.
However, it seems that the face attribute research community has so far relied informal definitions being sufficient.
At the other extreme, the two annotators had essentially random agreement on high cheekbones, pointy nose and oval face, disagreeing on 512, 490 and 466 of 1,000 images, respectively.
Based on these results, we divide the 40 attributes into three ranges of inherent consistency.
Attributes labeled consistently for $\geq 95\%$ of the 1,000 images have \textit{green} background in Table~\ref{table:consistency_1000}.
Attributes with consistency $>$ 85\% and $<$ 95\% have \textit{yellow} background and attributes with $\leq$ 85\%  consistently have \textit{pink} background. \textit{\textbf{The results show that the ambiguity of definition can bring a large inconsistency on annotations.}}

\subsection{Consistency Across Duplicate Faces?}
\begin{table*}[ht]
\centering
\begin{tabular}{|l|l|l|l|l||l|l|l|l|l|}
\hline
Attribute           & $\mathcal{N}_{differ}$ & $N_{n}$ & $N_{p}$ & $\mathcal{P}_{in}$  &  Attribute             & $\mathcal{N}_{differ}$ & $N_{n}$ & $N_{p}$ & $\mathcal{P}_{in}$  \\ \hline
Blurry             & 154 & 9,836 & 300   & 0.529 & 5 o Clock Shadow   & 192 & 8,616 & 1,520 & 0.149 \\ \hline
Pointy Nose       & 860 & 7,338 & 2,798 & 0.425 & Mustache              & 79  & 9,566 & 570   & 0.147 \\ \hline
Pale Skin         & 144 & 9,644 & 492   & 0.308 & Black Hair           & 270 & 7,644 & 2,492 & 0.144 \\ \hline
Rosy Cheeks       & 226 & 9,294 & 842   & 0.293 & High Cheekbones      & 343 & 5,396 & 4,740 & 0.136 \\ \hline
Oval Face         & 535 & 7,710 & 2,426 & 0.290 & Sideburns             & 90  & 9,418 & 718   & 0.135 \\ \hline
Narrow Eyes       & 280 & 8,936 & 1,200 & 0.265 & Goatee                & 79  & 9,484 & 652   & 0.130 \\ \hline
Straight Hair     & 447 & 7,920 & 2,216 & 0.258 & Big Lips             & 275 & 7,092 & 3,044 & 0.129 \\ \hline
Chubby             & 159 & 9,424 & 712   & 0.240 & Heavy Makeup         & 249 & 6,128 & 4,008 & 0.103 \\ \hline
Wearing Necklace  & 316 & 8,524 & 1,612 & 0.233 & Young                 & 190 & 2,640 & 7,496 & 0.097 \\ \hline
Bags Under Eyes  & 454 & 7,456 & 2,680 & 0.230 & Gray Hair            & 49  & 9,594 & 542   & 0.096 \\ \hline
Brown Hair        & 376 & 8,038 & 2,098 & 0.226 & Bald                  & 25  & 9,830 & 306   & 0.084 \\ \hline
Receding Hairline & 201 & 9,140 & 996   & 0.224 & Bangs                 & 111 & 8,560 & 1,576 & 0.083 \\ \hline
Double Chin       & 139 & 9,446 & 690   & 0.216 & Blond Hair           & 112 & 8,530 & 1,606 & 0.083 \\ \hline
Wearing Earrings  & 333 & 8,096 & 2,040 & 0.204 & Smiling               & 196 & 5,072 & 5,064 & 0.077 \\ \hline
Bushy Eyebrows    & 280 & 8,418 & 1,718 & 0.196 & No Beard             & 119 & 1,920 & 8,216 & 0.077 \\ \hline
Wavy Hair         & 441 & 6,700 & 3,436 & 0.194 & MSO & 178 & 5,214 & 4,922 & 0.070 \\ \hline
Attractive         & 482 & 4,486 & 5,650 & 0.193 & Wearing Lipstick     & 159 & 5,254 & 4,882 & 0.063 \\ \hline
Arched Eyebrows   & 430 & 6,758 & 3,378 & 0.191 & Wearing Hat          & 12  & 9,860 & 276   & 0.045 \\ \hline
Big Nose          & 343 & 6,956 & 3,180 & 0.157 & Eyeglasses            & 10  & 9,662 & 474   & 0.022 \\ \hline
Wearing Necktie   & 140 & 9,110 & 1,026 & 0.152 & Male                  & 12  & 5,648 & 4,488 & 0.005 \\ \hline
\end{tabular}
\caption{Level of inconsistency in CelebA attributes based on analysis of duplicate face appearances.}
\label{table:duplicate_pairs}
\end{table*}
Web-scraped datasets naturally contain duplicate and near-duplicate images.
The same face image may appear on multiple websites with different brightness or other edits; these are duplicates.
Also, images may be taken at slightly different times or from different points of view; these are near duplicates.
To identify duplicate face appearances in CelebA, we start with image pairs from the same identity group that have ArcFace (ResNet101 backbone) similarity $\geq 0.9$, and manually inspect these pairs to discard near-duplicates,
This resulted in 5,068 duplicate pairs from 3,094 identities.
Additional duplicates could be found by extending to lower ArcFace similarity, but 5K+ pairs is more than sufficient for a useful estimate of the consistency of CelebA attribute values.

If the CelebA attribute values are assigned in a perfectly consistent manner, the attribute values for each duplicate image pair should be identical.
An attribute whose values are assigned at random with a 50/50 true/false split is expected to have consistent values on 50\% of the duplicate pairs.
An attribute whose values are assigned at random with a 90/10 true/false split is expected to have consistent attribute values on 82\% (81\% + 1\%) of the duplicate pairs.
In general, the level of inconsistency observed in a particular attribute’s values across the duplicate pairs can be computed as:
\begin{equation}
    \mathcal{P}_{in}=\frac{\mathcal{N}_{differ}}{(\mathcal{P}(n|p)+\mathcal{P}(p|n))\times \mathcal{N}_{total}}
\end{equation}
where 
$\mathcal{N}_{total}$ is total number of duplicate image pairs, 
$\mathcal{N}_{differ}$ is number of duplicate pairs with attribute values that differ,
$\mathcal{P}(n|p) + \mathcal{P}(p|n)$ is the number of image pairs expected to agree if the values are assigned at random with the overall relative frequency.
The estimated inconsistency level ranges from 0, representing perfectly consistent, to 1, representing a level of inconsistency indicating values assigned at random.

Table~\ref{table:duplicate_pairs} summarizes, for each of the 40 attributes, the number of duplicate pairs whose attribute values differ, the number of total negative (attribute=false) samples, the number of total positive (attribute=true) samples, and the level of inconsistency computed using the above equation.
The attributes are listed from high to low level of inconsistency.
The highest level of inconsistency is for ``blurry'', which is actually an image attribute rather than a face attribute.
The same source image appearing online with, for example, two different levels of compression could  appear different in blur.
For this reason, we focus on the other 39 attributes for analyzing the consistency of attribute values.

There are 12 attributes with inconsistency less than 0.1, 14 with inconsistency between 0.1 and 0.2, and 14 with inconsistency above 0.2.
The facial attribute pointy-nose has the highest inconsistency, at 0.425 and the attribute male has the lowest, at 0.005, for a difference of two orders of magnitude.
\textit{\textbf{It is clear that the level of inherent inconsistency varies greatly across attributes for duplicate image pairs.}}

\subsection{Accuracy of CelebA Attribute Values?}
\begin{table}[!t]
\centering
\begin{tabular}{|l|l|l|l|l|}
\hline
Attribute & $N_n$ & $N_p$ & $Err_n$                          & $Err_p$                          \\ \hline
\cellcolor[HTML]{C6E0B4}Bald              & 972              & 500              & \cellcolor[HTML]{70AD47}0.41\%  & \cellcolor[HTML]{FF0000}43.40\% \\ \hline
\cellcolor[HTML]{C6E0B4}Eyeglasses        & 925              & 500              & \cellcolor[HTML]{70AD47}0.43\%  & \cellcolor[HTML]{70AD47}3.00\%  \\ \hline
\cellcolor[HTML]{C6E0B4}Goatee            & 938              & 500              & \cellcolor[HTML]{70AD47}0.96\%  & \cellcolor[HTML]{FF0000}57.40\% \\ \hline
\cellcolor[HTML]{C6E0B4}Gray Hair        & 953              & 500              & \cellcolor[HTML]{70AD47}3.78\%  & \cellcolor[HTML]{FF0000}17.00\% \\ \hline
\cellcolor[HTML]{C6E0B4}Male              & 563              & 500              & \cellcolor[HTML]{70AD47}0.18\%  & \cellcolor[HTML]{70AD47}4.20\%  \\ \hline
\cellcolor[HTML]{C6E0B4}MSO & 513 & 500 & \cellcolor[HTML]{FF0000}20.86\% & \cellcolor[HTML]{70AD47}1.60\%  \\ \hline
\cellcolor[HTML]{C6E0B4}No Beard         & 500              & 837              & \cellcolor[HTML]{FF0000}39.20\% & \cellcolor[HTML]{70AD47}1.19\%  \\ \hline
\cellcolor[HTML]{C6E0B4}Earrings$^*$ & 793              & 500              & 8.07\%                          & \cellcolor[HTML]{FF0000}17.00\% \\ \hline
\cellcolor[HTML]{C6E0B4}Hat$^*$      & 954              & 500              & \cellcolor[HTML]{70AD47}1.05\%  & 5.80\%                          \\ \hline
\cellcolor[HTML]{C6E0B4}Necklace$^*$ & 877              & 500              & 5.36\%                          & \cellcolor[HTML]{FF0000}55.80\% \\ \hline
\cellcolor[HTML]{C6E0B4}Necktie$^*$      & 928 & 500 & \cellcolor[HTML]{70AD47}0.97\%  & \cellcolor[HTML]{FF0000}30.00\% \\ \hline
\cellcolor[HTML]{C6E0B4}Mustache          & 959              & 500              & 7.09\%                          & 9.60\%                          \\ \hline
\end{tabular}
\caption{Error rate estimation of objective attributes. $N$ is the number of random samples. $n$ stands for negative in the original annotation, $p$ stands for positive in the original annotation. $*$ are the wearings.}
\label{table:error_rate_objective}
\end{table}

Above experiments show that some attributes can be annotated with high consistency across annotators and across different versions of the same image.
Restricting our attention to the 12 attributes with $\geq 95\%$ consistent manual annotations, we now estimate the error rate in the attribute values distributed with CelebA.

We increase the number of images annotated to a minimum of 500 with each original CelebA value for each of the 12 attributes.
Two persons independently assign values for each image.
For this experiment, attribute values that are initially inconsistent are considered again to arrive at a consensus correct value.
The consensus values are used to mark the original CelebA attribute values as correct or incorrect.
Table~\ref{table:error_rate_objective} summarizes the error rate, broken out for true/false original CelebA  value, and color-coded with green background for $\leq 5\%$ error and red for $> 15\%$ error.

Error rates for original CelebA attribute values vary greatly across attributes, and between true/false values for each attribute.
For instance, the error rate for (bald=true) is over 43\% and for (bald=false) it is only 0.41\%, for over two orders of magnitude difference.
High error rates and highly imbalanced error rates can present problems for machine learning algorithms.

\begin{figure*}[t]
\centering
  \includegraphics[width=\linewidth]{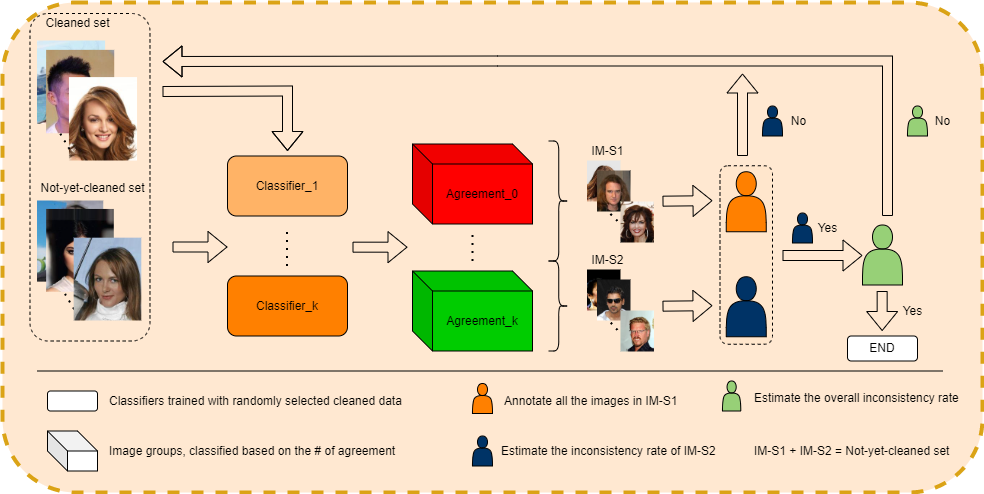}
  \caption{Quality-audit workflow for cleaning the annotations. Our annotation cleaning process is ended when the estimated inconsistency rate of all annotations is lower than 5\%.}
  \label{fig:mso_auditing_workflow}
\end{figure*}
Beyond plain error, attribute values in red can be categorized into three causes.
One, there is increased ambiguity for facial hair attributes such as goatee and no\_beard occur when hair length is short.
Two, ``wearing'' attributes (necklace, necktie, earrings) may have errors if the original values were assigned from viewing full images whereas the cleaned values were assigned from viewing cropped images. 
For example, a necklace that is visible in the full image may not be visible in the cropped image.
We are choosing for the ``correct'' value to be what is visible in the cropped face image.
Three, errors for other attributes (gray\_hair, MSO, bald) may arise from ambiguous definitions. 
(Examples of each type are shown in Figures 1, 2, and 3 of  the Supplementary Material).
The cause of errors is speculative, since, again, we do not know how the original CelebA attribute values were assigned.
In general, even for attributes that in principle can have consistent annotations, the error rate in the original CelebA annotations seems subjectively ``high'' for use as training data for machine learning models.

\section{Models From Cleaned Versus Original}

In principle, higher quality attribute values should enable learning a better model.
To investigate this, we select one attribute for cleaning/correction and compare the quality of the models learned from corrected versus original values.
We select the MSO attribute because the original values are relatively evenly distributed between true/false, its estimated error rates for true/false are representative of the asymmetric error rate problem, and corrected values can be assigned with high consistency.

\subsection{Audited Data-Clean Workflow}
\label{section:workflow}

To reduce the error rate of the 0.2M  MSO values, we propose an efficient quality-audit workflow, shown in Figure~\ref{fig:mso_auditing_workflow}. 
The workflow starts with a small subset of manually cleaned data, so that values are consistent and with low error.
Then, we randomly select $k$ subsets from this cleaned subset, and use each subset to train a classifier (i.e. $k$ trained classifiers in total).
The not-yet-cleaned data is classified by each of the $k$ classifiers, where the results are use to calculate the agreement level.
Based on agreement of the $k$ classifiers, the not-yet-cleaned data is divided into $k+1$ subsets.
The error rates in the subsets with high agreement between classifiers are estimated by a manual audit, and if it is below a target error rate, the subset is kept as passing the audit.
Subsets that have a low level of agreement among classifiers, but are small enough, are manually labeled.
Subsets that are too large to be manually labeled and whose estimated error rate exceeds the target are moved on to the next round
The workflow ends when the overall estimated error is below the target rate.

Table~\ref{table:error_rate_objective} estimates the MSO error rate as 20.9\% for negative (false) values and 1.6\% for positive.
A target error rate of below 5\% suggests that cleaning should focus on the images with original false values. 
For the initial subset of cleaned data, we manually labeled the images in the original test set, leaving the original train and validation sets (90\% of the data) untouched. \textit{Consequently, we manually cleaned 60,858 (30\%) of images from the original dataset.}

\begin{figure*}[t]
\centering
    \begin{subfigure}[b]{1\linewidth}
    \captionsetup[subfigure]{labelformat=empty}
    \centering
        \begin{subfigure}[b]{0.49\linewidth}
            \begin{subfigure}[b]{0.32\linewidth}
                \includegraphics[width=\linewidth]{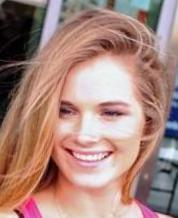}
            \end{subfigure}
            \begin{subfigure}[b]{0.32\linewidth}
                \includegraphics[width=\linewidth]{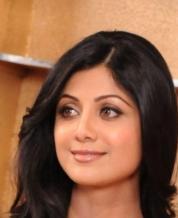}
            \end{subfigure}
            \begin{subfigure}[b]{0.32\linewidth}
                \includegraphics[width=\linewidth]{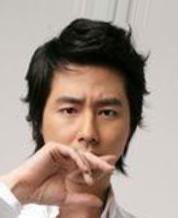}
            \end{subfigure}
        
        \caption{Open, Close, and Invisible MSO attribute}
        \label{fig:mso_samples-a}
        \end{subfigure}
        \begin{subfigure}[b]{0.49\linewidth}
             \begin{subfigure}[b]{0.32\linewidth}
                \includegraphics[width=\linewidth]{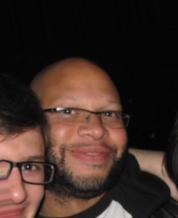}
            \end{subfigure}
            \begin{subfigure}[b]{0.32\linewidth}
                \includegraphics[width=\linewidth]{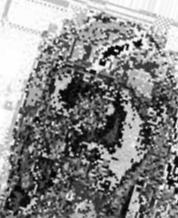}
            \end{subfigure}
            \begin{subfigure}[b]{0.32\linewidth}
                \includegraphics[width=\linewidth]{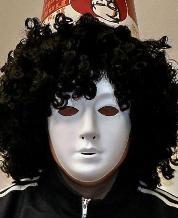}
            \end{subfigure}
        
        \caption{Unusable images}
        \label{fig:mso_samples-b}
        \end{subfigure}
        \begin{subfigure}[b]{0.49\linewidth}
            \begin{subfigure}[b]{0.32\linewidth}
                \includegraphics[width=\linewidth]{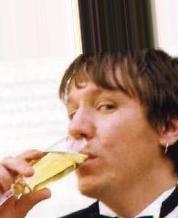}
            \end{subfigure}
            \begin{subfigure}[b]{0.32\linewidth}
                \includegraphics[width=\linewidth]{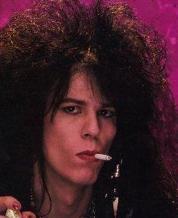}
            \end{subfigure}
            \begin{subfigure}[b]{0.32\linewidth}
                \includegraphics[width=\linewidth]{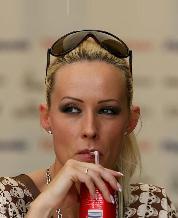}
            \end{subfigure}
        
        \caption{Ambiguous samples}
        \label{fig:mso_samples-c}
        \end{subfigure}
        \begin{subfigure}[b]{0.49\linewidth}
            \begin{subfigure}[b]{0.32\linewidth}
                \includegraphics[width=\linewidth]{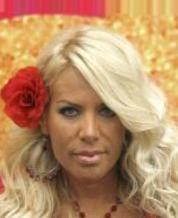}
            \end{subfigure}
            \begin{subfigure}[b]{0.32\linewidth}
                \includegraphics[width=\linewidth]{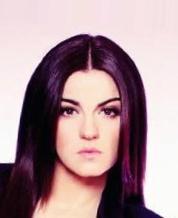}
            \end{subfigure}
            \begin{subfigure}[b]{0.32\linewidth}
                \includegraphics[width=\linewidth]{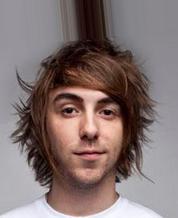}
            \end{subfigure}
        
        \caption{Edge images}
        \label{fig:mso_samples-d}
        \end{subfigure}
    \end{subfigure}
   \caption{Examples of three annotations options, unusable images, ambiguous images, and edge images of MSO attribute (More examples are shown in Figures 4, 5, 6 from the supplemental material).}
\label{fig:mso_samples}
\end{figure*}
\begin{table*}[t]
\centering
\begin{tabular}{|l|l|l|l|l|l||l|l|}
\hline
         & Train   & Val    & Test   & Info\_not\_vis & Unusable & NMSO     & MSO    \\ \hline
Original & 162,770 & 19,867 & 19,962 & $ - $         & $ - $       & 104,657 (51.7\%) & 97,942 (48.3\%) \\ \hline
Cleaned  & 161,982 & 19,741 & 19,913 & 797       & 166      & 73,704 (36.6\%)  & 127,932 (63.4\%) \\ \hline
\end{tabular}
\caption{Comparison between original dataset and cleaned dataset on both annotation and image. Info\_not\_vis means the information of mouth is not visible in the image.}
\label{table:dataset_comparison}
\end{table*}
\begin{table}[t]
\centering
\begin{tabular}{|l|l|l|l|l|}
\hline
Train/Val/Test             & AFF       & MOON        & RN50     & DN121  \\ \hline
Or/Or/Or & 94.16          & 94.09          & 93.95          & 94.10          \\ \hline
Or/Or/Cl  & 85.17          & 85.94          & 85.24          & 85.98          \\ \hline
Or/Cl/Cl   & 86.17          & 86.49          & 86.54          & 85.69          \\ \hline
Cl/Cl/Or   & 86.13          & 85.40          & 85.90          & 85.27          \\ \hline
Cl/Cl/Cl    & \textbf{95.18} & \textbf{95.49} & \textbf{95.33} & \textbf{95.15} \\ \hline
\end{tabular}
\caption{Model accuracy on different combinations of original and cleaned MSO values used for train/validation/test set. Or = original, Cl = cleaned, AFF = AFFACT, RN = ResNet, DN = DenseNet.}
\label{table:network_performances}
\end{table}

\subsection{Definition of MSO}
\label{section:definition_mso}

Figure~\ref{fig:mso_samples} shows problems enountered in a detailed definition of MSO.
A small number of images do not contain a visible human face, and so are dropped.
A slightly larger but still small number of images do not contain visible information to assign a value for some attributes, and so we introduce a third possible attribute value, \textit{information not visible}. For example, if a person is holding a microphone in front of their face, the information may not be visible to assign attribute values for mouth, lips or nose.

To assign MSO values consistently, we make a detailed definition, including some edge cases.
For example, in Figure~\ref{fig:mso_samples-c}, there is something - e.g. glass, cigarette, straw - between the lips but the other part of lips is touching.
For our definition, these edge cases are defined as part of (MSO=true).
However, it is inevitable that there are still edge cases where it is uncertain if the lips are touching or not (see Figure~\ref{fig:mso_samples-d}).
These cases should dominate the images with label ``errors'' in the audited/cleaned attributes.

Table~\ref{table:dataset_comparison} summarizes the original and cleaned dataset.
The cleaned dataset has 166 images dropped as unusable for facial attribute classification, and an additional 797 images with (MSO=info\_not\_visible). Note that most of the images in these two categories are not in the test set.
A substantial fraction of the images have not been manually examined in our quality-audit workflow,
so it is possible that more images that could be dropped and more attribute values could be (MSO=info\_not\_visible) if all images were manually reviewed.
The MSO-cleaned dataset has 161,982 training samples, 19,741 validation samples, and 19,913 testing samples. 
The number of positive and negative samples are not as balanced as in the original dataset.

\subsection{Higher Quality, Higher Accuracy}
\label{section:network performance}

To investigate the impact of learning models using the cleaned attribute values, we train four algorithms with different combinations of cleaned versus original values.
The four algorithms are AFFACT~\cite{gunther2017affact} (implementation obtained from original authors), MOON~\cite{rudd2016moon} (re-implemented), ResNet50~\cite{he2016deep} (from Pytorch~\cite{paszke2019pytorch}) and DenseNet121~\cite{huang2017densely} (also PyTorch).
We are experimenting with one attribute, MSO, so the last fully connected layer goes to one output.
Input images are resized to $224\times224$. MSE loss is used for training MOON, and binary cross-entropy loss for the other three. Each is trained for 50 epochs with 128 batch size. The learning rate is 0.00001 for MOON, and 0.0001 for the others. Horizontal flipping is used for data augmentation.

Table~\ref{table:network_performances} summarizes the accuracies obtained by the four algorithms.
For train/val/test with all original data, AFFACT obtains the highest accuracy, 94.16\%.
This is in keeping with what would be expected from the literature \cite{gunther2017affact, thom2020facial}.
For train/val/test with clean data, each algorithm obtains higher accuracy than with original data, {\it all four obtain an accuracy that exceeds the state-of-the-art obtained for this attribute with the original annotations}, and MOON obtains the highest, 95.49\%.

\begin{figure*}[t]
    \centering
    \begin{subfigure}[b]{1\linewidth}
    \captionsetup[subfigure]{labelformat=empty}
        \begin{subfigure}[b]{1\linewidth}
            \begin{subfigure}[b]{0.32\linewidth}
                \begin{subfigure}[b]{0.49\linewidth}
                    \includegraphics[width=\linewidth]{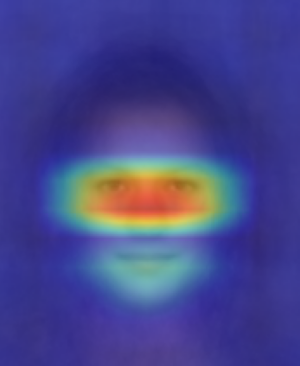}
                \end{subfigure}
                \begin{subfigure}[b]{0.49\linewidth}
                    \includegraphics[width=\linewidth]{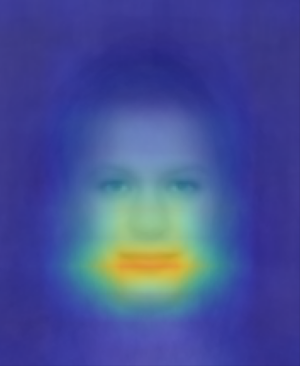}
                \end{subfigure}
            \caption{MSO=False}
            \end{subfigure}
            \begin{subfigure}[b]{0.32\linewidth}
                \begin{subfigure}[b]{0.49\linewidth}
                    \includegraphics[width=\linewidth]{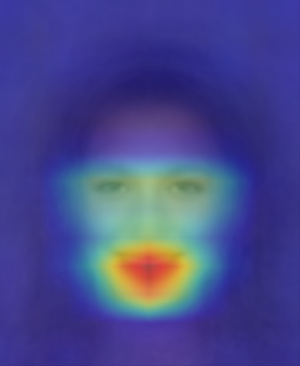}
                \end{subfigure}
                \begin{subfigure}[b]{0.49\linewidth}
                    \includegraphics[width=\linewidth]{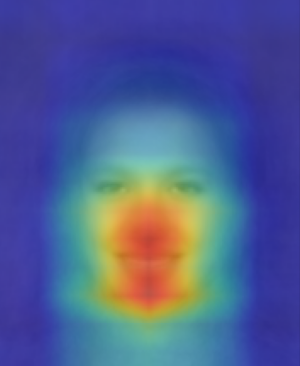}
                \end{subfigure}
            \caption{MSO=True}
            \end{subfigure}
            \begin{subfigure}[b]{0.32\linewidth}
                \begin{subfigure}[b]{0.49\linewidth}
                    \includegraphics[width=\linewidth]{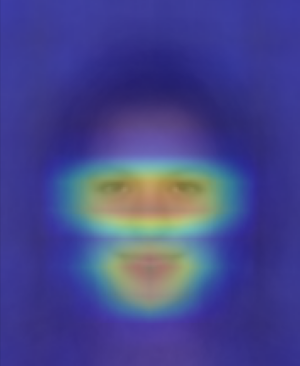}
                \end{subfigure}
                \begin{subfigure}[b]{0.49\linewidth}
                    \includegraphics[width=\linewidth]{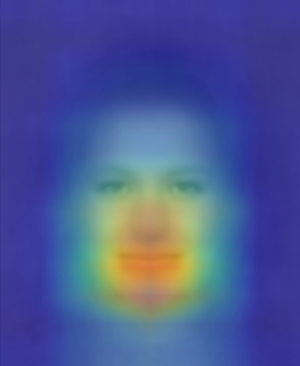}
                \end{subfigure}
            \caption{Overall}
            \end{subfigure}
        \end{subfigure}
    \end{subfigure}
   \caption{Average Score-CAM of the MOON created using original annotations and our cleaned annotations. The left side images of each pair is the average CAM generated by MOON trained with original dataset. The right side image of each pair is the average CAM generated by MOON trained with cleaned dataset.}
\label{fig:average-cams}
\end{figure*}

The various combinations of mixed original and cleaned for train/val/test all suffer a dramatic drop in accuracy.
This supports the premise that higher quality training data enables better models, and partial measures of cleaning only the test data, or only the validation and test, are counter-productive.
If the collective values of the training data are not a quality approximation to the real-world concept, then the networks learn a model of some different concept.
The starkness of the difference can be appreciated from the Score-CAM\cite{wang2020score, jacobgilpytorchcam}  visualization of the average heat maps of the models.
Figure~\ref{fig:average-cams} compares the average Score-CAM heatmap for MOON across the test set images for the model learned using our cleaned attribute values and the original attribute values.
Face location is normalized in CelebA images, so the mouth appears on average in the same area of the image.
The heatmap for the model learned using our cleaned  values is clearly focused on the mouth region, as it should be to judge mouth open/closed.
(The examples of heatmap comparison of individuals can be found in Figure 7 from the supplementary material). 
The heatmap for the model learned from original attribute values has no discernible focus on the mouth region.
This is visual evidence that the models learned from the cleaned and the original data are fundamentally different, and that the model learned from the cleaned data is better focused on the desired real-world concept.

\section{Discussion and Conclusions}

We seek to better understand the quality of models learned for facial attribute classification.
We begin with an experiment to assess the consistency of human annotations for the set of 40 commonly-used binary facial attributes.
We find that only 12 of the 40 are readily annotated with $\geq$ 95\% consistency between annotators.
This result (a) highlights that most research datasets are created and distributed with no assessment of the consistency or reproducibility of the ``ground truth labels'', (b) points to the need for more careful curation of meta-data, and (c) suggests that some face attributed may be too subjective to be useful.

Another experiment focuses specifically on the consistency of attribute values distributed with CelebA, the most widely used research dataset in this area.
We identify 5,068 image pairs in CelebA with duplicate facial appearances, compare the  attribute values across these pairs, and find that every attribute has contradicting values for some duplicate pairs, 
with the level of disagreement ranging up to 860 of the 5,068 pairs (for ``pointed nose'').

The level of attribute value disagreement on CelebA duplicate face appearances motivates a more careful assessment of the correctness of the CelebA attribute values.
This assessment focuses on the 12 attributes that we found can be annotated with with $\geq 95\%$ consistency.
Estimated error rates are highly asymmetric between the two values of an attribute; for example, the error rate for (MSO=false) is estimated at 20.9\% and for (MSO=true) at 1.6\%.
Asymmetric error rates this high raise questions of what concept a model learns from this training data.

To investigate further, we create a cleaned version of the CelebA attribute values for MSO.
The first step is manually-assigned correct MSO values for a small subset of CelebA.
We use the initial subset of manually-assigned correct MSO values to bootstrap a hybrid automated/manual process that results in a cleaned version of the CelebA MSO values.
(This cleaned version of the CelebA MSO values will be made available with the final paper.)
The intended level of quality in our cleaned version of the CelebA MSO values is that a new independent manual labeling of the MSO attribute would agree at $\geq 95\%$ level with for both the positive and negative values of the attribute.
As a side effect of creating the cleaned version of the CelebA MSO attribute, a small number of original CelebA images are marked as unusable in general due to not containing a human face image, and a small number are marked as ``info not visible'' due to the mouth being occluded in the image.

To assess the impact of our cleaned attribute values for learning a model, we compare accuracy of four algorithms trained using our cleaned versus the original attribute values.
For all four algorithms, the model learned using cleaned values achieves higher accuracy than the state-of-the-art for models using the original data.
Our implementation of the MOON algorithm achieves 95.49\% accuracy using the cleaned dataset.
Even more than the difference in accuracy levels achieved, comparing the Score-CAM heatmaps convincingly shows that the model learned using cleaned data is fundamentally better aligned with the attribute being classified.

Our results suggest that facial attribute classification research will benefit from datasets with higher quality meta-data.
At the least, research datasets should come with a description of the level accuracy that can be expected in its attribute values.
It may also be important, for multi-attribute classification, to allow for different subsets of attributes to be used from different images, in order to accommodate occlusion of different parts of the face in different images.
The CelebA dataset has been foundational for research in face attribute classification, and now potentially spurs new directions.

\section{Acknowledgement}

This research is based upon work supported in part by the Office of the Director of National Intelligence (ODNI), Intelligence Advanced Research Projects Activity (IARPA), via \textit{2022-21102100003}. The views and conclusions contained herein are those of the authors and should not be interpreted as necessarily representing the official policies, either expressed or implied, of ODNI, IARPA, or the U.S. Government. The U.S. Government is authorized to reproduce and distribute reprints for governmental purposes notwithstanding any copyright annotation therein.

{\small
\bibliographystyle{ieee_fullname}
\bibliography{egbib}
}

\end{document}